\documentclass{article} 
\pdfoutput=1 
\usepackage{iclr2020_conference,times}

\usepackage{amsmath,amsfonts,bm}

\def\eqref#1{equation~\ref{#1}}

\def\1{\bm{1}}

\DeclareMathAlphabet{\mathsfit}{\encodingdefault}{\sfdefault}{m}{sl}
\SetMathAlphabet{\mathsfit}{bold}{\encodingdefault}{\sfdefault}{bx}{n}

\usepackage{url}
\usepackage{graphicx}
\usepackage{booktabs}
\usepackage{cite}
\usepackage{hyperref}
\usepackage{subcaption}
\usepackage{geometry}

\title{Masakhane - Machine Translation For Africa}

\author{$\forall$, Iroro Fred \d{\`O}n\d{\`o}m\d{\`e} Orife,
Julia Kreutzer,
Blessing Sibanda,
Daniel Whitenack,
Kathleen Siminyu,
Laura Martinus,
Jamiil Toure Ali,
Jade Abbott,
Vukosi Marivate,
Salomon Kabongo,
Musie Meressa,
Espoir Murhabazi,
Orevaoghene Ahia,
Elan van Biljon,
Arshath Ramkilowan,
Adewale Akinfaderin,
Alp Öktem,
Wole Akin,
Ghollah Kioko,
Kevin Degila,
Herman Kamper,
Bonaventure Dossou,
Chris Emezue,
Kelechi Ogueji,
Abdallah Bashir\thanks{In rebellion against the status attributed to author order, the order of authors has been randomised~\citep{strange2008}. The symbol $\forall$ furthermore takes the place as first author to represent the whole community.}\\
Masakhane, Africa\\
masakhane.io \\
\texttt{masakhane-mt@googlegroups.com} \\
}

\author{\\\parbox{\textwidth}{\textbf{
$\forall$, Iroro Fred \d{\`O}n\d{\`o}m\d{\`e} Orife,
Julia Kreutzer,
Blessing Sibanda,
Daniel Whitenack,
Kathleen Siminyu,
Laura Martinus,
Jamiil Toure Ali,
Jade Abbott,
Vukosi Marivate,
Salomon Kabongo,
Musie Meressa,
Espoir Murhabazi,
Orevaoghene Ahia,
Elan van Biljon,
Arshath Ramkilowan,
Adewale Akinfaderin,
Alp Öktem,
Wole Akin,
Ghollah Kioko,
Kevin Degila,
Herman Kamper,
Bonaventure Dossou,
Chris Emezue,
Kelechi Ogueji,
Abdallah Bashir}}\thanks{In rebellion against the status attributed to author order, the order of authors has been randomised~\citep{strange2008}. The symbol $\forall$ furthermore takes the place as first author to represent the whole community.}\\
Masakhane, Africa\\
masakhane.io \\
\texttt{masakhane-mt@googlegroups.com} \\
}

\iclrfinalcopy 
\begin{document}

\maketitle

%
\section{The State of African NLP}\label{sec:motivation}


2144 of all 7111 (30.15\%) living languages today are African languages~\citep{enthnologue}. 
But only a small portion of linguistic resources for NLP research are built for African languages. As a result, there are only few NLP publications:
 In all ACL conferences in 2019, only 5 out of 2695 (0.19\%) author affiliations were based in Africa \citep{caines_2019}. This stark contrast of \emph{linguistic richness versus poor representation} of African languages in NLP is caused by multiple factors.

First of all, African societies do not see hope for African languages being accepted as primary means of communication~\citep{alexander2009evolving}. As a result, few efforts to fund NLP or translation for African languages exist, despite the potential impact. This \emph{lack of focus} has had a ripple effect.
 
The few existing resources are not easily discoverable, published in closed journals, non-indexed local conferences, or remain undigitized, surviving only in private collections~\citep{mesthrie1995language}. This \emph{opaqueness} impedes researchers' ability to reproduce and build upon existing results, and to develop, compete on and progress public benchmarks~\citep{martinus2019focus}. 

African researchers are disproportionately affected by \emph{socio-economic factors}, and are often hindered by visa issues~\citep{johnson_2019} and costs of flights from and within Africa~\citep{hattem_hattem_2017}. They are distributed and disconnected on the continent, and rarely have the opportunity to commune, collaborate and share.

Furthermore, African languages are of \emph{high linguistic complexity and variety}, with diverse morphologies and phonologies, including lexical and grammatical tonal patterns, and many are practiced within multilingual societies with frequent code switching \citep{ndubuisi2019wetin, bird1999strategies, gibbon2006morphotonology}. Because of this complexity, cross-lingual generalization from success in languages like English are not guaranteed.


\section{Contribution}
Founded at the \textit{Deep Learning Indaba 2019},
\textsc{Masakhane} constitutes an open-source, continent-wide, distributed, online research effort for machine translation for African languages.
Its goals are threefold:


\begin{enumerate}
    \item \textbf{For Africa}: To build a community of NLP researchers, connect and grow it, spurring and sharing further research, to enable language preservation and increase its global visibility and relevance.
    \item \textbf{For NLP researchers}: To build data sets and tools to facilitate NLP research on African languages, and to pose new research problems to enrich the NLP research landscape.
    \item \textbf{For the global researchers community}: To discover best practices for distributed research, to be applied by other emerging research communities.
\end{enumerate}

\section{Methodology and Results}\label{sec:methods}

\begin{figure*}[t]
    \centering
      \begin{subfigure}[b]{0.4\textwidth}
    \includegraphics[width=\columnwidth]{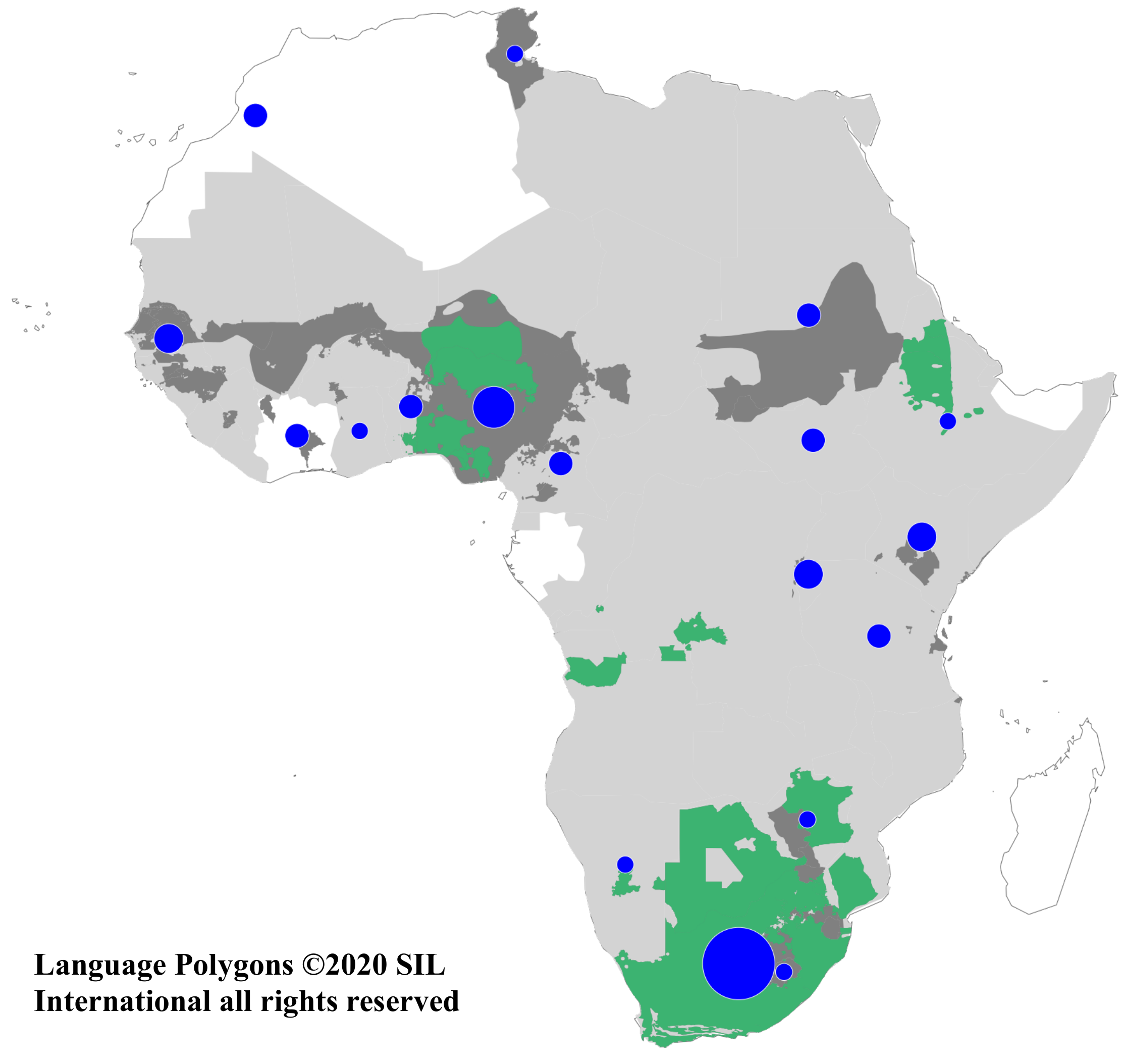}
    \caption{Origin of Participants \& Focus Language.} 
    \label{fig:map}
    \end{subfigure}
    \hspace{1cm}
    \begin{subfigure}[b]{0.3\textwidth}
    \centering
    \includegraphics[width=\columnwidth]{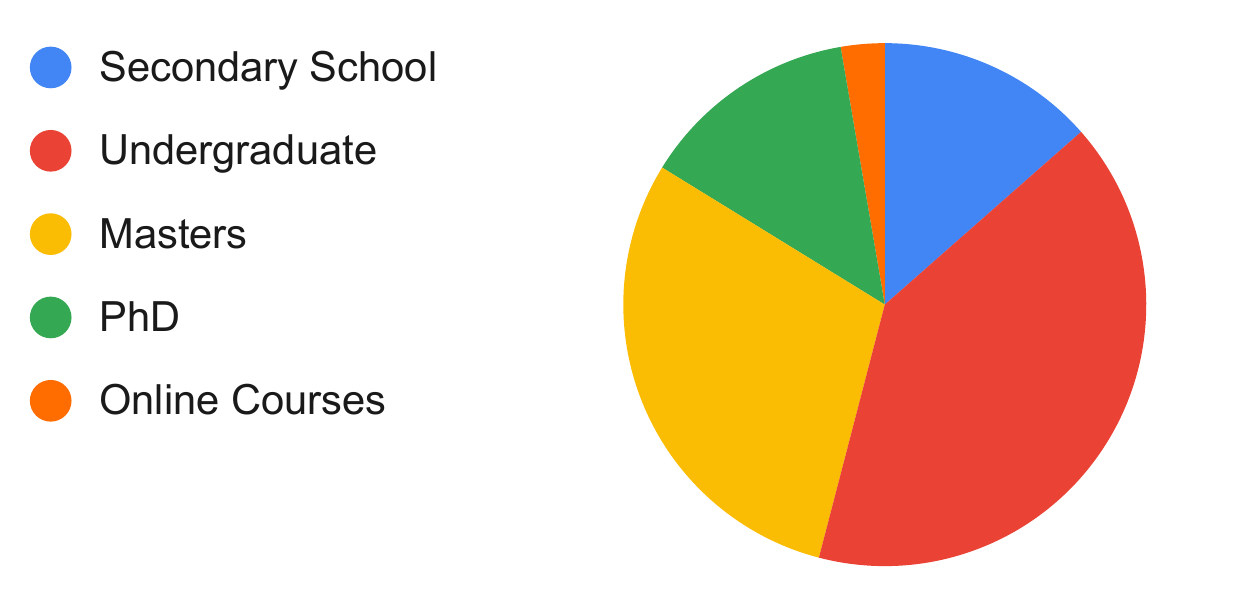}
    \caption{Highest Level of Education.}
    \label{fig:education}
    \includegraphics[width=\columnwidth]{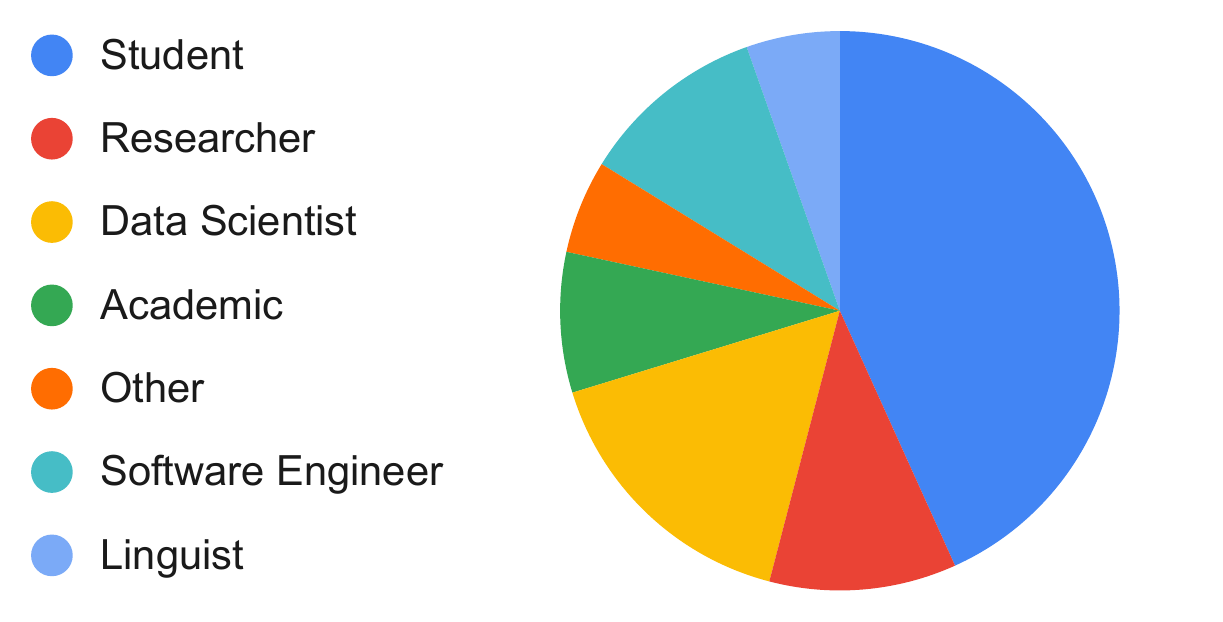}
    \caption{Occupation.}
    \label{fig:occupation}
    \end{subfigure}
    \caption{Participants with African origin are represented by blue markers in (a), indigenous areas that are covered by the languages of current benchmarks in green, benchmarks in progress in dark grey, and countries where those languages are spoken in light grey. Education (b) and occupation (c) of a subset of 37 participants as indicated in a voluntary survey in February 2019.}
    \label{fig:demographics}
\end{figure*}

\textsc{Masakhane}'s strategy is to offer \emph{barrier-free open access} to first hands-on NLP experiences with African languages, fighting the above-mentioned opaqueness. With an easy-to-use open source platform, it allows individuals to train neural machine translation (NMT) models on a parallel corpus for a language of their choice, and share the results with an online community. The \emph{online community} is based on weekly meetings, an active Slack workspace, and a GitHub repository (\texttt{github.com/masakhane-io}),
so that members can support each other and connect despite geographical distances. \emph{No academic prerequisites} are required for participation, since tertiary education enrolments are minimal in sub-saharan Africa \citep{britishcouncil}. 

A \emph{Jupyter Notebook} features documented data preparation, model configuration, training and evaluation. It runs on Google Colab with a single (free) GPU for a small limited number of hours, such that participants do not require expensive hardware. The NMT models are built using Joey NMT \citep{joey2019}, which comes with a beginner-friendly documentation. Participants submit and publish their data, code and results for training on their language to improve reproducibility and discoverability. 
To lower the barrier of data collection, the JW300 multilingual dataset \citep{agic-vulic-2019-jw300} with parallel corpora for English to 101 African languages is integrated into the notebook. With the goal of improving translation quality by transfer learning across languages in the future, \emph{global test sets} with English sources are extracted from JW300, and excluded from training data for any language pair to avoid potential data leakage for cross-lingual transfer.

As of February 14, 2020, the \textsc{Masakhane} community consists of 144 participants from 17 African countries with diverse educations and occupations (Figure~\ref{fig:demographics}),
and 2 countries outside Africa (USA and Germany).
So far, 30 translation results for 28 African languages have been published by 25 contributors on GitHub.

\section{Future Roadmap}
\textsc{Masakhane} aims to continue to grow and facilitate engagement within the community, especially helping inactive users contribute benchmarks and fostering mentoring relations. In the next year, the project will expand to different NLP tasks beyond NMT in order to reach a broader audience. Qualitative analysis on model performance as well as investigations of automatic evaluation metrics will spur healthy competition on results. \textsc{Masakhane} will also provide notebooks for transfer and un/self-supervised learning to push translation quality. In terms of data collection, the size and domain of global test sets will be expanded.

\bibliography{iclr2020_conference}
\bibliographystyle{iclr2020_conference}

\end{document}